\documentclass[a4paper,fleqn]{cas-dc}

\usepackage[numbers]{natbib}

\usepackage{amsfonts}
\usepackage{amsmath}
\usepackage{amsthm}
\usepackage{graphicx}
\usepackage{multirow}
\usepackage{url}
\usepackage{xcolor}
\usepackage{booktabs}
\usepackage{subfigure}
\def\tsc#1{\csdef{#1}{\textsc{\lowercase{#1}}\xspace}}
\tsc{WGM}
\tsc{QE}
\tsc{EP}
\tsc{PMS}
\tsc{BEC}
\tsc{DE}

\begin{document}
\shorttitle{Designing Deep Networks for Scene Recognition}
\shortauthors{Zhinan Qiao, Xiaohui Yuan}

\title [mode = title]{Designing Deep Networks for Scene Recognition}                      


\author[1]{Zhinan Qiao}

\author[1]{Xiaohui Yuan}


\address[1]{University of North Texas, 1155 Union Cir, Denton, TX 76203}









\begin{abstract}
Most deep learning backbones are evaluated on ImageNet. Using scenery images as an example, we conducted extensive experiments to demonstrate the widely accepted principles in network design may result in dramatic performance differences when the data is altered. Exploratory experiments are engaged to explain the underlining cause of the differences. Based on our observation, this paper presents a novel network design methodology: data-oriented network design. In other words, instead of designing universal backbones, the scheming of the networks should treat the characteristics of data as a crucial component. We further proposed a Deep-Narrow Network and Dilated Pooling module, which improved the scene recognition performance using less than half of the computational resources compared to the benchmark network architecture ResNets. The source code is publicly available on https://github.com/ZN-Qiao/Deep-Narrow-Network.
\end{abstract}




\maketitle

\section{Introduction}
Since the development of AlexNet~\cite{krizhevsky2012imagenet}, a number of variations of deep Convolutional Neural Networks (CNNs) emerged. Motivated by the success of studies~\cite{simon14, szege16, 16He}, the networks become deeper by adding more convolutional layers to improve performance for the targeted problems. On the other hand, arguments on the benefits of increasing network width gained support from researchers. Zagoruyko et al.~\cite{zagoruyko2016wide} presented wider deep residual networks that significantly improve the performance over ResNet~\cite{16He}. The wider deep network achieved state-of-the-art performance on ImageNet~\cite{deng09} and CIFAR~\cite{krizhevsky2009learning}, showing consistently better accuracy than its ResNet counterparts. Xie et al.~\cite{xie17} proposed a multi-branch architecture called ResNeXt by widening the residual blocks and cooperating group convolution. In contrast to ResNet, ResNeXt~\cite{xie17} demonstrated a performance boost on a larger ImageNet-5K set and the COCO object detection dataset~\cite{lin2014microsoft}. ResNeSt~\cite{zhang2020resnest} preserved the wider network layout and multi-branch strategy and introduced a modulated architecture to improve the feature learning process. The proposed ResNeSt networks further improved the performance on the ImageNet dataset.

Despite the benefits brought by increased depth and width (i.e., number of channels), side effects such as the requirement of large number of training examples make it difficult to employ a large-scale network for learning from complex scenery images. Is a deeper network more suitable for extracting features from scenery images for better understanding complex views? Or shall we employ more channels to achieve improved performance? To answer these questions, we need to have a better understanding of the functions of network layers and channels. A number of studies have been conducted in the past years. Lu et al.~\cite{lu2017expressive} argued that an integration of both depth and width provides a better understanding of the expressive power of neural networks. Tan and Le~\cite{tan2019efficientnet} showed that it is critical to balance the network depth and width by maintaining a constant depth/width ratio and demonstrated the effectiveness of their approach on ResNet and MobileNet. Besides manually designed networks, deep Neural Architecture Search were proposed to optimize the network depth and width~\cite{zoph2016neural,guo2020dmcp}. However, most methods, if not all, were developed and evaluated using ImageNet~\cite{deng09} and CIFAR~\cite{krizhevsky2009learning}. These datasets commonly depict an object near the center of the image, and the label tells what the object is. That is, such an image is mostly dominated by one object and, hence, is referred to as ``object-centric". On the other hand, a scene image presents a complex view that consists of multiple objects and background clutters. This inadvertent data bias could potentially lead to the ignorance of the characteristics of different data. The features that are crucial for object recognition dominantly affect the design of the CNNs. 

Deep networks with more layers use a variety of receptive fields to extract distinctive scale features whereas networks with more channels capture fine-grained patterns~\cite{tan2019efficientnet}. However, in many applications, both types of information are crucial for accurately recognizing an image,  but the prominence difference between them is seldom explored. 

This paper attempts to bring out a new perspective on neural network design: the data itself has the preference. Learning the overall spatial layout is crucial to recognize the entire scene, thus scene recognition favors the networks that can better learn the spatial information. For the object-centric images, typical examples only consist of one single object, the spatial layout does not contribute much to the semantic meaning of the image. As the differences between certain object categories are subtle, the detailed patterns and textures of objects are likely to be more representative. The networks that emphasize learning various features can better fit the requirement of object recognition tasks. By considering the distinct characteristics of scene images, our hypothesis is that for scene recognition, learning spatial-wise information improves the performance of CNNs in a more effective manner compared to learning channel-wise information. To evaluate this hypothesis, we conducted comprehensive experiments and our results show scene images gain clear benefits from deepening the network, and the performance change caused by altering the width is marginal. We further proposed a Deep-Narrow Network, which increases the depth of the network as well as decreased the width of the network. We design a Dilated Pooling component and use it in our Deep-Narrow Network to extract spatial features. Our design presents an efficient approach to scale the network based on the data. 

We summarize our key contributions as follows:
\begin{enumerate}
\item We argue that the primary factor in designing neural networks should be the data being used, rather than designing universal network architectures. Learning the spatial layout is important for scene recognition tasks, while learning various features is more beneficial for object recognition tasks. We conducted experiments to validate our hypothesis.

\item We analyze the effect of depth and width in CNNs from the perspective of image characteristics and demonstrate that, compared to wider networks, deeper networks are more suitable for performing scene recognition tasks. To prove this hypothesis, we conducted extensive experiments to analyze the underlying factors affecting the performance. Our findings suggest that spatial information is crucial for scene recognition. Networks that can learn spatial information effectively perform better in recognizing the overall spatial layout of a scene. This is in contrast to object-centric images, where the detailed patterns and textures of objects are more representative.

\item Based on our hypothesis, we proposed the Deep-Narrow Network, an efficient approach to scale the network by increasing its depth while decreasing its width. This design includes a Dilated Pooling component to extract spatial features. Our experimental results show that deepening the network provides clear benefits for scene images, with marginal performance changes caused by altering the width. This approach provides an efficient way to scale the network based on the data characteristics.

\end{enumerate}

\section{Related Work}
\textbf{Deep Networks}
Network depth has played an integral role in the success of CNNs. With the increase of depth, the network can better approximate the target function with richer feature hierarchies, which enables the boost of performance. The success of VGGNet~\cite{simon14} and Inception~\cite{szege15,szege16,szege16a} on ILSVRC competition further reinforced the significance of the depth. ResNet ~\cite{he16}, which is a continuance work of deeper networks, revolutionized the possible depth of deep networks by introducing the concept of residual learning and identity mapping into CNNs. ResNet's effective methodology enables the network to be extremely deep and demonstrated improved performance in image recognition tasks.

\textbf{Wide Networks}
Network width has also been suggested as an essential parameter in the design of deep networks. Wide ResNet~\cite{zagoruyko2016wide} introduced an additional factor to control the width of the ResNet. The experimental results showed that the widening of the network might provide a more effective way to improve performance compared to making ResNet deeper. Xception~\cite{choll17} can be considered as an extreme Inception architecture, which exploits the idea of depth-wise separable convolution. Xception modified the original inception block by making it wider. This wider structure has also demonstrated improved performance. ResNeXt~\cite{xie17} introduced a new term: cardinality to increase the width of ResNet and won the 2016 ILSVRC classification task. With the success of ResNeXt, it is widely accepted that widening the deep network is an effective way to boost model performance. A most recent wider network: ResNeSt~\cite{zhang2020resnest} preserved the wide architecture of ResNeXt and achieved superior performance on image and object recognition tasks.

\textbf{Effects of Depth and Width}
Although depth and width are proven to be essential parameters in network architecture design, the effect of depth and width, i.e., what do deep and wide networks learn remains seldom explored. Most of the existing literature focus on the effect of width and depth separately or the trade-off between depth and width in the network design~\cite{lu2017expressive}. Tan and Le~\cite{tan2019efficientnet} claimed that deep networks can make use of a larger receptive field while wide networks can better capture fine-grained features. Nguyen et al.~\cite{nguyen2020wide} explored the effects of width and depth and found a characteristic structure named block structure. They demonstrated that for different models, the block structure is unique, but the representations outside the block structure trends to be similar despite the setting of depth and width. In our paper, we analyze the effect of depth and width in CNNs from the perspective of image characteristics.

\textbf{Scene Recognition}
Integrating multiple networks serves as the pioneer scene recognition study in the era of deep learning and is widely adopted in the community~\cite{zhu19, cheng18, qiao2020urban}. Network outputs are concatenated or integrated together to generate a uniform scene feature representation and demonstrated improved performance~\cite{herra16, wang17b, deng09, zhou16, cheng18, xia19, selva17, qiao2020urban}. In these designs, pre-trained networks are typically used as feature extractors and the output will be sent into a classifier to make final decision. Although in some studies, trainable integration methods have been devised~\cite{seong19, seong19a}, the backbone network architectures remain unchanged.

Besides network integration, there exist other methods that use different algorithms and network architectures to address the scene recognition challenge~\cite{shi2019scene, wang2020deep, rehman2021scene, gupta2021visual}. However, in these studies, although different scene recognition components are leveraged,
the network backbones are still used as feature extractors. To the best of our knowledge, in most recent works, add-on modules are introduced in scene recognition and achieved desirable results~\cite{qiao2021attention, yuan2021scale}, but the network backbone design remains seldom explored. Apparently, there exists a gap between state-of-the-art generic deep neural network design and scene recognition network design. Introducing the most advanced network design theories and practices into the scene recognition domain is imminent and imperative.

\section{Experimental Study}
\subsection{Data Sets and Experimental Settings}
To understand the impact of network structure on the dataset and ultimately the applications, we use ImageNet 2012~\cite{deng09} and Places Standard dataset~\cite{zhou17} as our evaluation datasets. ImageNet 2012 is the benchmark object recognition dataset that consists of 1,000 classes and 1.28 million training images. An image in ImageNet 2012 usually contains a single object that is highly distinctive from the background. Place365 Standard dataset is designed for scene recognition and contains 1.8 million training images of 365 classes. The images in Places365 Standard datasets present more complex scenery images.

We train deep network models and compute the single-crop ($224\times224$ pixels) top-1 and top-5 accuracy based on the application of the models to the validation set. We train each model for 100 epochs on eight Tesla V100 GPUs with 32 images per GPU (the batch size is 256). All models are trained using synchronous SGD (Stochastic Gradient Descent) with a Nesterov momentum of 0.9 and a weight decay of 0.0001. The learning rate is set to 0.1 and is reduced by a factor of 10 in every 30 epochs. In the training of ResNet and its variants, we follow the settings in~\cite{he16}.

\subsection{Comparison on Different Data Sets}
We conducted our comparison study using ResNet and its variants on Places365 and ImageNet datasets. The results are reported in Tables~\ref{tb:comparision1} (varying network depth) and~\ref{tb:comparision2} (varying network width). By increasing the network depth from 50 to 101, i.e., ResNet-50 and ResNet-101, we obtained a performance improvement of 1.40\% and 2.32\% on Place365 and ImageNet datasets in terms of Top-1 accuracy. Theoretically, if the widening of the network is more effective to improve the performance as stated in the previous literature, we should expect more accuracy increase when we double the width of the networks. However, doubling the width leads to a top-1 accuracy increase of 3.28\% on ImageNet, but only 0.94\% on Place365. More surprisingly, for ResNeXt, which also doubled the network width, the relative performance increase on ImageNet is 2.34\% in terms of top-1 accuracy, but the number is only 0.14\% on Place365. 


\begin{table}[!htb] 
\caption{Top-1 and top-5 accuracy (\%) comparison by changing the network depth.
\label{tb:comparision1}}
\centering
\setlength{\tabcolsep}{3mm}{
\begin{tabular}{c|l|cc}
\toprule
Data &\multicolumn{1}{c}{Model} & Top-1 & Top-5\\
\midrule\midrule
& ResNet-18  & 54.22 & 84.63 \\ 
Places365 & ResNet-50 & 55.69 & 85.80 \\
& ResNet-101 & 56.47 & 86.25 \\
\midrule
& ResNet-18 & 70.52 & 89.56 \\ 
ImageNet & ResNet50 & 76.02 & 92.80 \\
& ResNet-101 & 77.78 & 93.72 \\
\bottomrule
\end{tabular}
}
\end{table}


\begin{table}[!htb] 
\caption{Top-1 and top-5 accuracy (\%) comparison by changing the network width. The numbers within parenthesis are the width scaling factors.
\label{tb:comparision2}}
\centering
\setlength{\tabcolsep}{3mm}{
\begin{tabular}{c|l|cc}
\toprule
Data &\multicolumn{1}{c}{Model} & Top-1 & Top-5 \\
\midrule\midrule
&ResNet-50 ($\times$ 1) & 55.69 & 85.80 \\ 
&ResNet-50 ($\times$ 2) & 56.21 & 86.11 \\
Places365 & ResNet-50 ($\times$ .5) & 55.07 & 85.12 \\
&ResNet-50 ($\times$ .25) & 52.16 & 82.85 \\
&ResNeXt-50 & 55.77 & 85.99 \\
\midrule
&ResNet-50 ($\times$ 1) & 76.02 & 92.80 \\ 
&ResNet-50 ($\times$ 2) & 78.51 & 94.09 \\
ImageNet&ResNet-50 ($\times$ .5) & 72.08 & 90.78 \\
&ResNet-50 ($\times$ .25) & 64.04 & 85.76 \\ 
&ResNeXt-50 & 77.80 & 94.30 \\
\bottomrule
\end{tabular}
}
\end{table}

The trend that the model performance on ImageNet is more sensitive to width change compared with Place365 is also true when we decrease network depth or narrow down the width. When the network depth decreased from 50 to 18, ImageNet suffered a 7.23\% relative top-1 accuracy decrease, and for Place365 it is 2.71\%. That is, the top-1 performance drop on ImageNet is around 2.7 times of the top-1 performance drop on Places. But when we decreased the width of ResNet-50 to half of the original size, the number changed to 4.7. The statement that widening of the network might provide a more effective way to improve performance is biased towards ImageNet (object-centric data) and ignores the characteristics of scenery images.

\begin{table}[!htb] 
\caption{Top-1 and top-5 accuracy (\%) comparison by changing the network depth using 100 randomly selected classes.
\label{tb:comparision100_1}}
\centering
\setlength{\tabcolsep}{3mm}{
\begin{tabular}{c|l|cc}
\toprule
Data &\multicolumn{1}{c}{Model} & Top-1 & Top-5\\
\midrule\midrule
Places365 & ResNet-18  & 70.63 & 94.32 \\ 
& ResNet-50 & 71.14 & 94.56 \\
\midrule
ImageNet & ResNet-18 & 81.17 & 94.50 \\ 
& ResNet50 & 83.19 & 95.42 \\
\bottomrule
\end{tabular}
}
\end{table}

\begin{table}[!htb] 
\caption{Top-1 and top-5 accuracy (\%) comparison by changing the network width using 100 randomly selected classes. The numbers within parenthesis are the width scaling factors.
\label{tb:comparision100_2}}
\centering
\setlength{\tabcolsep}{3mm}{
\begin{tabular}{c|l|cc}
\toprule
Data &\multicolumn{1}{c}{Model} & Top-1 & Top-5 \\
\midrule\midrule
& ResNet-50 ($\times$ 1) & 71.14 & 94.56 \\ 
Places365 & ResNet-50 ($\times$ .5) & 71.04 & 94.21 \\
&ResNet-50 ($\times$ .25) & 70.14 & 93.76 \\
\midrule
&ResNet-50 ($\times$ 1) & 83.19 & 95.42 \\ 
ImageNet&ResNet-50 ($\times$ .5) & 80.17 & 94.06 \\
&ResNet-50 ($\times$ .25) & 76.93 & 92.62 \\ 
\bottomrule
\end{tabular}
}
\end{table}

As Places365 has 365 classes while ImageNet has 1000 classes, to avoid the bias caused by the different numbers of classes between the two datasets, we randomly selected 100 classes from each dataset and conducted comparison experiments. As we shrunk the size of the datasets, we observed under-fitting occurred when we apply the smaller dataset to large models. So we only include the comparison results on smaller models in Table~\ref{tb:comparision100_1} and Table~\ref{tb:comparision100_2}. We observe that for the Places dataset, switching ResNet-50 to ResNet-18 leads to a 0.72\% top-1 accuracy drop on Places and 2.49\% top-1 accuracy drop on ImageNet, ie the performance drop on ImageNet is around 3.46 times of the drop on Places. Meanwhile, narrowing the width of ResNet-50 to 1/2 and 1/4 of the original width leads to 0.14\% and 1.41\% top-1 performance drop, for ImageNet, the accuracy decrease is 3.6\% and 7.5 \%. The performance drop caused by halving the width on ImageNet is 25.7 times the performance drop on Places, which verified that our hypothesis that altering network width has a less significant effect on scenery data is not biased towards the number of classes in the dataset.

\section{Analysis of the Impact of Depth and Width of Deep Networks}
\subsection{Complexity of Images}
\begin{figure*}[!htb]
\centering
\begin{tabular}{cccc}
\includegraphics[width=1.55in]{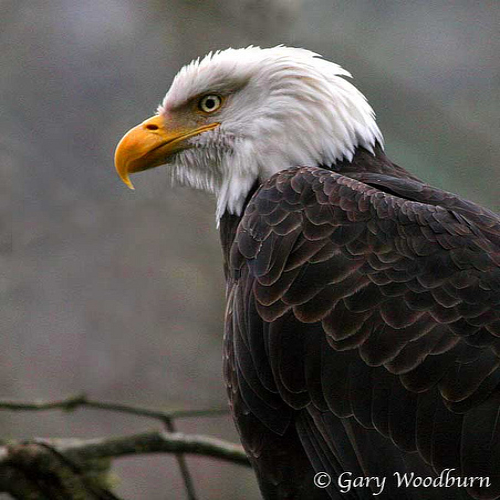} &
\includegraphics[width=1.55in]{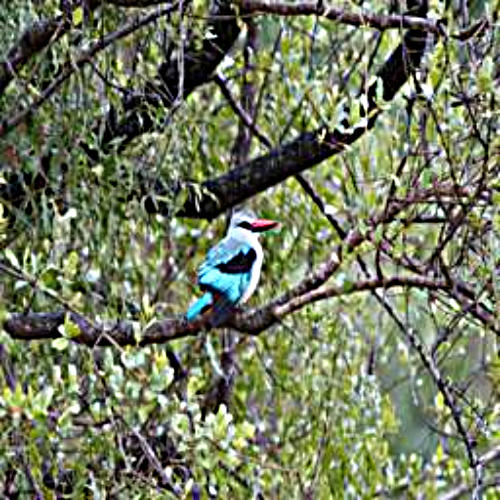} &
\includegraphics[width=1.55in]{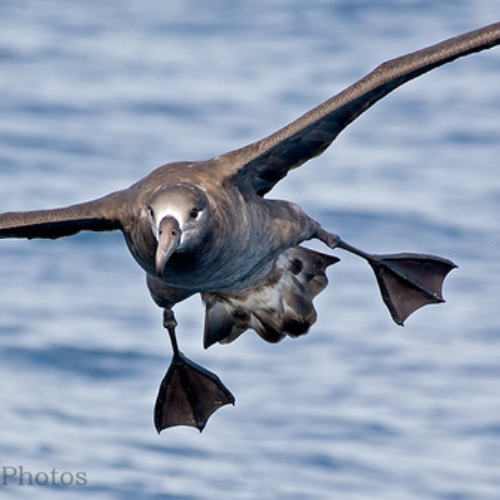} &
\includegraphics[width=1.55in]{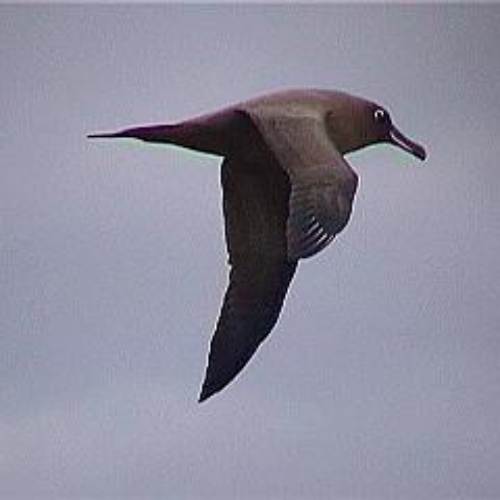}\\
(a) ImageNet & (b) Place365 & (c) CUB-200-2011 & (d) CUB-200-2011
\\ (Bald eagle)  & (Forest-broad leaf) & (Black-footed albatross) & (Sooty albatross)
\end{tabular}
\caption{Example images from benchmark object recognition dataset ImageNet (a), benchmark scene recognition dataset Places365 (b), and fine-grained dataset Caltech-UCSD Birds-200-2011 (c and d).
\label{fig:imagenet} }
\end{figure*}

Our first hypothesis is the performance difference is caused by the complexity of scenery images. This hypothesis is aroused by the distinct complexity difference between scene images and object-centric images: object-centric images always only contain one major object which occupies a large portion of the view, scene images always consist of multiple objects and background clutters. Figure~\ref{fig:imagenet} shows two samples from benchmark object-centric dataset (ImageNet) and benchmark scene dataset (Place365), respectively. Figure~\ref{fig:imagenet} (a) is labeled as ``bald eagle", in which the eagle stands in the center of the view and occupies a large portion of the entire image; figure~\ref{fig:imagenet} (b) is labeled as ``forest-broad leaf", the entire view consists of not only a bird but also tree branches and leaves. As the correct recognition of scenery images relies on multiple components, a scene image is typically considered more complex than an object-centric image.

To evaluate this hypothesis, we introduced another dataset that is also widely believed to be ``complex": the fine-grained image classification dataset. Fine-grained classification is considered to be a more complex task as the classes in the dataset can only be discriminated by local and subtle differences. CUB-200-2011 is a dataset consisting of 200 different species of birds, which serves as a benchmark dataset for fine-grained classification tasks. Figure~\ref{fig:imagenet} shows two samples in the CUB-200-2011 dataset. In Figure~\ref{fig:imagenet}, black-footed albatross (c) and sooty albatross (d) are considered to be two different categories in classification. The two albatrosses are similar in appearance, and the differentiating of them is challenging due to the subtle traits that characterize the different species are not straightforward. 

We conducted comparison experiments on two ``complex" datasets (Places365 and CUB-200-2011) and one ``simple" dataset. The results are shown in Table~\ref{tb:fine-grained}. Using the benchmark ResNet-50 as the backbones, we observed that on CUB-200-2011, the relative top -1 accuracy increased by 1.81\% when doubling the width and dropped 4.28\% when we narrowed the width to half of the original. This performance change caused by altering the width is much acute compared to the result on Place365 (0.94\% and 1.11\%, doubling and halving the width respectively) under the same settings, which demonstrated that a wider network is able to effectively enhance the recognition of ``complex" fine-grained features. This result does not agree with our first hypothesis: if the performance difference is originated from the complexity of the data, we should observe a moderate performance change on CUB-200-2011 dataset along with the changing of network width. The results proved that the performance variance on different dataset is not the consequence of the complexity (rich fine-grained details) of the data. 

\begin{table}[!htb]
\caption{Top-1 accuracy (\%) of ResNet-50 on Places365, ImageNet, and Caltech-UCSD Birds-200-2011 by changing network width.
\label{tb:fine-grained}}
\centering
{\small
\begin{tabular}{c|ccc}
\toprule
Width Scaling Factor & Places & ImageNet & CUB\\ 
\midrule\midrule
2 & 56.21 & 78.51 & 71.53 \\ 
1 & 55.69 & 76.02 & 70.26 \\
0.5 & 55.07 & 72.08 & 67.25 \\
0.25 & 52.16 & 64.04 & 61.29 \\
\bottomrule
\end{tabular}
}
\end{table}

\subsection{Spatial v.s. Channel}
Based on the observations, we came up with the second hypothesis: for the scene recognition task, instead of learning more fine-grained features, learning spatial information is more crucial. As defined in~\cite{fan2020analyzing}, spatial information refers to the spatial ordering on the feature map. Intuitively, for images that only contain one object, the semantic meaning related to spatial layout is limited; for scene images, the spatial structures, namely, scene contextual information likely to contribute more to the understanding of the scene. Thus, for scene recognition tasks, learning spatial information is more crucial. 

\begin{figure*}[!htb]
\begin{center}
\begin{tabular}{cc}
\includegraphics[width=5.3in]{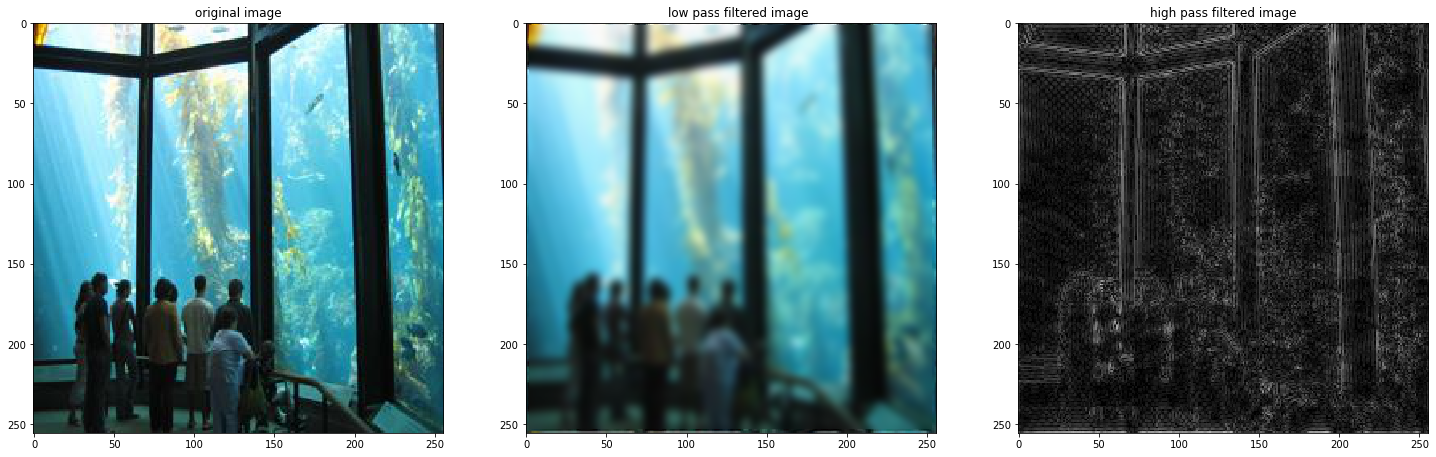}
\end{tabular}
\caption{Results on scenery image using low and high pass filters.
\label{fig:fftr}}
\end{center}
\end{figure*}

To verify this hypothesis, we conducted experiments by gradually passing low and high-frequency information on Place365 and ImageNet datasets. Generally speaking, the high-frequency information in the image refers to the regions where the intensity of the image (brightness/gray-scale) changes drastically, which are often called the edges or contour; the low-frequency information in the image refers to the regions where the image intensity changes smoothly, such as large patches of color. As shown in Figure~\ref{fig:fftr}, the image filtered by low-pass filters tends to present proximate or blurred patterns of the original image, the images filtered by high-pass filters better preserved the spatial information. 

\begin{figure*}[!htb]
\centering
\begin{tabular}{ccc}
\includegraphics[width=1.6in]{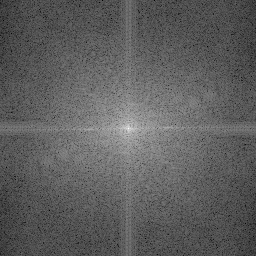} &
\includegraphics[width=1.6in]{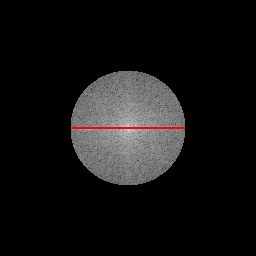} &
\includegraphics[width=1.6in]{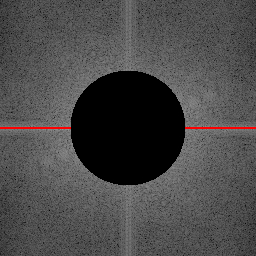} \\
(a) Image in frequency domain & (b) Low pass filter & (c) High pass filter
\end{tabular}
\caption{Illustration of low and high pass filters. In Figure (b) and (c), the mask (black region) denotes information removed by corresponding filters; the length of the red lines denotes the corresponding filter size. \label{fig:fft} }
\end{figure*}

To understand the importance of low and high-frequency information in different datasets, we designed the low-pass and high-pass filter based on Fourier Transform. We transformed the testing images into the spectrum domain using Fourier Transform and applied both low and high-pass filters to test how low/high-frequency information can affect the model performance on different datasets. Figure~\ref{fig:fft} shows the design of the filters: for low-pass filters, we masked the high-frequency components; and for high-pass filters, we masked the low-frequency components.

\begin{figure*}[!htb]
\centering
\begin{tabular}{cc}
\includegraphics[width=5.3in]{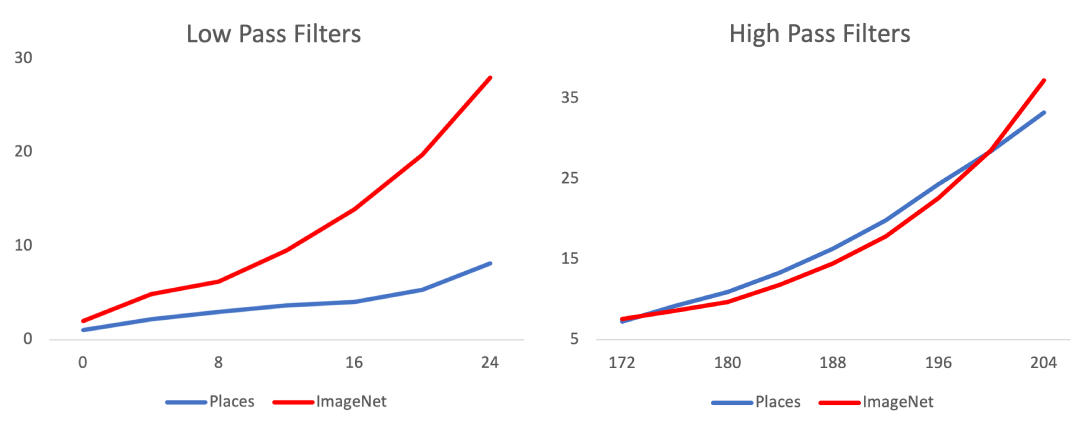}
\end{tabular}
\caption{Top-1 accuracy (\%) on Place365 and ImageNet datasets using low pass filters (left) and high pass filters (right). The x-axis denotes the size of the corresponding filters. \label{fig:filter} }
\end{figure*}

For a fair comparison, we randomly selected 100 classes from Place365 and ImageNet datasets to deploy the experiments. Note that as scene recognition is considered a harder task compared to object recognition, the classification accuracy on Place365 is lower despite the chance is the same.  The results are shown in Figure~\ref{fig:filter}. In both of the sub-figures, the x-axis denotes the size of the corresponding low/high pass filter in the spectrum domain (the maximum size is 224), and the y-axis denotes the top-1 accuracy. Through the comparison, we observed enlightening phenomena:  when we gradually pass the low frequency information, the performance increase on ImageNet is steeper than on Place365 (Figure~\ref{fig:filter} (a)). Surprisingly, when we use the low pass filter of size 33, ImageNet can achieve a top-1 accuracy of nearly 30\% while the chance is 1\%. This denotes the correct recognition of object-centric images heavily rely on low frequency information. On the contrary, when we gradually pass the high frequency information, we can observe that the model trained on scenery dataset is more sensitive to high-frequency information (Figure~\ref{fig:filter} (b)). Notably, when the size of the high-pass filter is around 210 to 214, the top-1 accuracy on Place365 even exceeds the top-1 accuracy on ImageNet despite the classification accuracy on ImageNet is always much higher on Place365 across various networks. This observation demonstrated that recognition of scenery images is susceptible to high frequency information.

Our observation perfectly fits the experimental results we discussed in Section 3. The high and low frequency information in images can approximately present the learned spatial and channel-wise information in deep networks. Wider networks have an expanded number of channels, which enables the network to learn more fine-grained features. Deep networks have an increasing number of layers and larger receptive fields, which enables the network to learn more spatial information. Thus for the scene recognition tasks, deepening the network is more efficient than increasing the width of the network. We proposed two directions to better enhance the scene recognition networks: deepening and narrowing the network, and decreasing the spatial information lost.

\section{Deep-Narrow Network}
\subsection{Deep-Narrow Structure}
\begin{figure*}[!htb] 
  \centering 
  \includegraphics[width=0.9\textwidth]{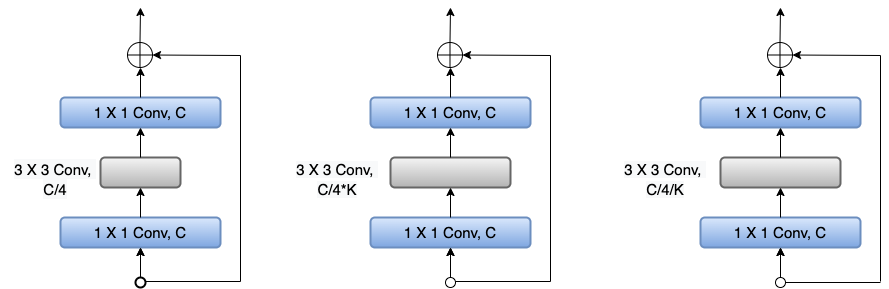} 
\caption{The schema of standard ResNet (left), Wide ResNet (middle), and Deep-Narrow Network (right). C denotes the number of channels. \label{fig:dn}} 
\end{figure*}

\begin{table*}[!htb]
\caption{Top-1 and Top-5 accuracy (\%) using ResNet with different depth and width. The number of parameters is in million. The numbers within parenthesis are the width scaling factors.
\label{tb:dn}}
\centering
\setlength{\tabcolsep}{5mm}{
\begin{tabular}{l|l|cccc}
\toprule
Data &Model & Top-1 & Top-5 & GFLOPs & $\#$ Parameters \\ 
\midrule\midrule
&ResNet-50 ($\times$ 1) & 55.69 & 85.80 & 4.12 & 24.26 \\
Place365 & ResNet-50 ($\times$ .5) & 55.07 & 85.12 & 1.07 & 6.27 \\ 
& Deep-Narrow Network & 55.58  & 85.80 & 2.00 & 11.03 \\
\midrule
&ResNet-50 ($\times$ 1) & 76.02 & 92.80 & 4.12 & 25.56 \\ 
ImageNet & ResNet-50 ($\times$ .5) & 72.08 & 90.78 & 1.07 & 6.92 \\
& Deep-Narrow Network & 74.99 & 92.31 & 2.00 & 11.68 \\
\bottomrule
\end{tabular}}
\end{table*}

Based on our observation, we argued that as correctly recognizing scenery images is susceptible to the learning of spatial information, designing the networks with larger depth and smaller width can potentially be an effective and efficient option. As shown in Figure~\ref{fig:dn}, we proposed a Deep-Narrow architecture. Different from the famous Wide ResNet~\cite{zagoruyko2016wide} which use the factor K to increase the width of the ResNet, we increase the number of layers in ResNet and decreases the width of the network using width factor K. In our designed Deep-Narrow Network, we use K equals to 2, i.e., we halve the network width used in benchmark ResNet.

Table~\ref{tb:dn} shows the performance comparisons among the benchmark ResNet-50, i.e., ResNet-50 ($\times$ 1), ResNet-50 with half the width, i.e., ResNet-50 ($\times$ .5)), and our Deep-Narrow Network on Place365 dataset. Deep-Narrow Network achieves comparable evaluation scores with benchmark ResNet-50 using only less than half of the FLOPs and parameters: the relative top-1 accuracy dropped 0.20\%. On the ImageNet, the Deep-Narrow Architecture obtained a relative top-1 accuracy drop of 1.35\%. The results re-certified that scene recognition is highly dependent on learning spatial information. Different from the assertions in the previous literature that widening of the network might provide a more effective way than by making ResNet deeper to improve the model performance, we demonstrated that data have their preference and the network design should rely on the characteristics of data.

\subsection{Dilated Pooling}
\begin{figure*}[!htb]
\centering
\begin{tabular}{c}
\includegraphics[width=4.3in]{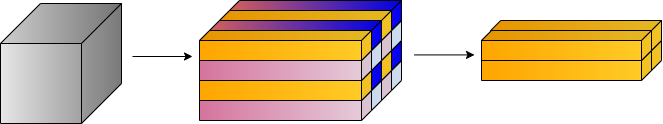} \\
(a) Down-sample component in ResNet, during the down-sampling process, 3/4 of the spatial in- \\ formation are discarded and the number of channels is quadrupled.
\\\\
\includegraphics[width=4.3in]{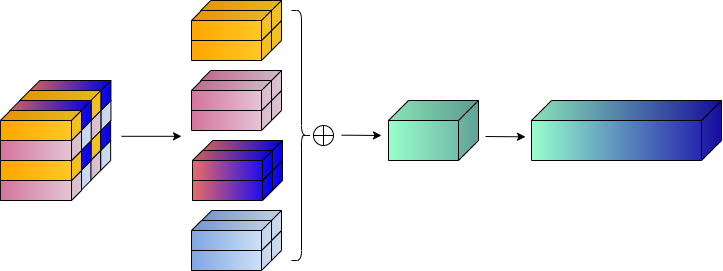} \\
(b) Dilated Pooling module which does not discard spatial information. 
\end{tabular}
\caption{The schema of down-sample component in ResNet and Dilated Pooling module. \label{fig:Dilated}}
\end{figure*}

\begin{figure*}[!htb]
\centering
\begin{tabular}{cp{0.3in}c}
\includegraphics[width=2in]{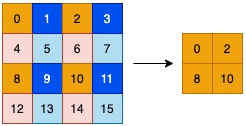} & &
\includegraphics[width=2in]{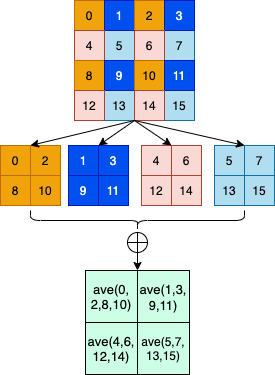} \\
(a) && (b)
\end{tabular}
\caption{The schema of the down-sample component in ResNet and  Pooling module from the view of spatial dimension. (a) Down-sample component in ResNet from the view of spatial dimension. During the down-sampling process, 3/4 of the spatial information are discarded and the number of channels is doubled. (b)  Pooling module from the view of spatial dimension. \label{fig:Dilated_2} }
\end{figure*}

Besides deepening the network to better process spatial information, we also designed a Dilated Pooling module to better preserve the spatial information in ResNet. As shown is Figure~\ref{fig:Dilated} (a), in ResNet, the down-sampling process is conducted by quadrupling the width of the network and discard 3/4 of the features along spatial dimension. Based on our statement, this design is not suitable for scene recognition as discarding spatial information is likely to devastate the performance. Alternatively, we designed a Dilated Pooling module to better preserve the spatial information (Figure~\ref{fig:Dilated} (b)). More details are illustrated in Figure~\ref{fig:Dilated_2}. In Dilated Pooling module, instead of directly discarding 3/4 of the spatial information, we separate the features maps into four sub-sections along the spatial dimension. Then we conduct convolution on the four feature maps and merged the result together via summation operation. By leveraging Dilated Pooling, we are able to make use of all the spatial information without increasing the Flops and number of parameters.

\begin{table*}[!htb]
\caption{Top-1 and top-5 accuracy (\%)  using Deep-Narrow Network and Dilated Pooling (DP) module. The number of parameters is in million.
\label{tb:Dilated_2}}
\centering
\setlength{\tabcolsep}{4mm}{
\begin{tabular}{l|l|cccc}
\toprule
Data & Model & Top-1 & top-5 & GFLOPs & Params \\ 
\midrule\midrule
& ResNet-50 ($\times$ 1) & 55.69 & 85.80 & 4.12 & 24.26 \\
Place365 & Deep-Narrow Network & 55.58  & 85.80 & 2.00 & 11.03 \\
& Deep-Narrow Network (with DP) & 55.91 & 86.12 & 2.00 & 11.03 \\ 
\midrule
& ResNet-50 ($\times$ 1) & 76.02 & 92.80 & 4.12 & 25.56 \\
ImageNet & Deep-Narrow Network & 74.99 & 92.31 & 2.00 & 11.68\\
& Deep-Narrow Network (with DP) & 74.63 & 92.13 & 2.00 & 11.68 \\
\bottomrule
\end{tabular}}
\end{table*}

\begin{table*}[!htb]
\caption{Top-1 and top-5 accuracy (\%) on Place365 using different backbones. The number of parameters is in million. Note that we calculate the number of parameters based on 365-class models (Place365). The best results are highlighted in bold.
\label{tb:last_compare}}
\centering
\setlength{\tabcolsep}{6mm}{
\begin{tabular}{l|cccc}
\toprule
Model & Top-1 & Top-5 & GFLOPs & $\#$ Parameter \\
\midrule\midrule
ResNet-50 ($\times$ 1) & 55.69 & 85.80 & 4.12 & 24.26 \\
Deep-Narrow Network & 55.58  & 85.80 & 2.00 & 11.03 \\
ResNet-Ave & 55.60 & 85.58 & 2.26 & 11.03 \\
ResNet-Max & 55.55 & 85.56 & 2.26 & 11.03 \\
ResNet-D & 55.79 & 85.93 & 2.26 & 11.03 \\
Antialiased-CNN & 55.85 & 85.93 & 2.26 & 11.03 \\
Deep-Narrow Network (with DP) & \textbf{55.91} & \textbf{86.12} & \textbf{2.00} & \textbf{11.03} \\
\bottomrule
\end{tabular}}
\end{table*}

As shown in Table~\ref{tb:Dilated_2}, by integrating the Dilated Pooling module with our Deep-Narrow Network design, we achieved better performance than the benchmark ResNet on the Place365 dataset, using less than half of the FLOPs and number of parameters. Specifically, adding Dilated Pooling to our Deep-Narrow Network resulted in a relative top-1 accuracy increase of 0.59\% (Table~\ref{tb:Dilated_2}), while only causing a 0.48\% relative top-1 accuracy drop on ImageNet. This demonstrates the effectiveness and efficiency of our data-oriented network design approach and the importance of designing networks according to the characteristics of the data.

To demonstrate the superiority of the proposed method, we compare with several state-of-the-art approaches that also tried to minimize the information lost caused by shrinking the spatial resolution, including ResNet-D~\cite{he2019bag} and Antialiased-CNN~\cite{zhang2019making}. ResNet-D used average pooling and convolution operation to maintain more spatial information; Antialiased-CNN leveraged convolution-based pooling strategies to enhance the conventional pooling. We also implemented two baseline strategies named ResNet-Ave and ResNet-Max, which simply use averaging or maxing operation instead of discarding 3/4 information. As shown in Table~\ref{tb:last_compare}, our model use 0.26 less GLOPs compared to the aforementioned methods, which is an 11.5\% computational resource saving. In short, we can observe that Deep-Narrow Network with Dilated Pooling achieves higher accuracy than the comparison methods using less computational resources.

\section{Conclusion}
This paper studies the impacts of scenery images in the network architecture design using ImageNet (object recognition dataset) and Places365 (scene recognition dataset) as examples. Carefully designed experiments showed that the characteristics of datasets affect the performance of the models: wider networks often achieve better performance of recognition of images with a prominent object and have less impact on the recognition of scenery images. We further evaluated our hypothesis by conducting comparison experiments and demonstrated that learning spatial-wise information is more substantial in scene recognition tasks compared to object classification tasks. This phenomenon explained why widening the networks is less effective than deepening the networks for scene recognition as deeper networks can better learn the spatial information in the training examples. Thus deploying the networks that have larger depth and smaller width, and emphasizing the spatial information learning likely to benefit scene recognition backbone designs. Our proposed Deep-Narrow Network and Dilated Pooling module re-certified the effectiveness and efficiency of taking advantage of data properly. Our design achieved a better accuracy compared to benchmark ResNet-50 using less than half of the computation resources.

\bibliographystyle{IEEEtran}
\bibliography{macros,main}

\begin{thebibliography}{10}
\providecommand{\url}[1]{#1}
\csname url@samestyle\endcsname
\providecommand{\newblock}{\relax}
\providecommand{\bibinfo}[2]{#2}
\providecommand{\BIBentrySTDinterwordspacing}{\spaceskip=0pt\relax}
\providecommand{\BIBentryALTinterwordstretchfactor}{4}
\providecommand{\BIBentryALTinterwordspacing}{\spaceskip=\fontdimen2\font plus
\BIBentryALTinterwordstretchfactor\fontdimen3\font minus
  \fontdimen4\font\relax}
\providecommand{\BIBforeignlanguage}[2]{{%
\expandafter\ifx\csname l@#1\endcsname\relax
\typeout{** WARNING: IEEEtran.bst: No hyphenation pattern has been}%
\typeout{** loaded for the language `#1'. Using the pattern for}%
\typeout{** the default language instead.}%
\else
\language=\csname l@#1\endcsname
\fi
#2}}
\providecommand{\BIBdecl}{\relax}
\BIBdecl

\bibitem{krizhevsky2012imagenet}
A.~Krizhevsky, I.~Sutskever, and G.~E. Hinton, ``Imagenet classification with
  deep convolutional neural networks,'' \emph{NIPs}, vol.~25, pp. 1097--1105,
  2012.

\bibitem{simon14}
K.~Simonyan and A.~Zisserman, ``Very deep convolutional networks for
  large-scale image recognition,'' \emph{ICLR}, 2015.

\bibitem{szege16}
C.~Szegedy, V.~Vanhoucke, S.~Ioffe, J.~Shlens, and Z.~Wojna, ``Rethinking the
  inception architecture for computer vision,'' in \emph{CVPR}, 2016, pp.
  2818--2826.

\bibitem{16He}
K.~He, X.~Zhang, S.~Ren, and J.~Sun, ``Deep residual learning for image
  recognition,'' in \emph{CVPR}, Los Alamitos, CA, USA, Jun 2016, pp. 770--778.

\bibitem{zagoruyko2016wide}
S.~Zagoruyko and N.~Komodakis, ``Wide residual networks,'' in
  \emph{BMVC}.\hskip 1em plus 0.5em minus 0.4em\relax British Machine Vision
  Association, 2016.

\bibitem{deng09}
J.~Deng, W.~Dong, R.~Socher, L.-J. Li, K.~Li, and L.~Fei-Fei, ``{ImageNet}: {A}
  large-scale hierarchical image database,'' in \emph{CVPR}, 2009, pp.
  248--255.

\bibitem{krizhevsky2009learning}
A.~Krizhevsky, G.~Hinton \emph{et~al.}, ``Learning multiple layers of features
  from tiny images,'' 2009.

\bibitem{xie17}
S.~Xie, R.~Girshick, P.~Doll{\'a}r, Z.~Tu, and K.~He, ``Aggregated residual
  transformations for deep neural networks,'' in \emph{CVPR}, 2017, pp.
  1492--1500.

\bibitem{lin2014microsoft}
T.-Y. Lin, M.~Maire, S.~Belongie, J.~Hays, P.~Perona, D.~Ramanan,
  P.~Doll{\'a}r, and C.~L. Zitnick, ``Microsoft coco: Common objects in
  context,'' in \emph{ECCV}.\hskip 1em plus 0.5em minus 0.4em\relax Springer,
  2014, pp. 740--755.

\bibitem{zhang2020resnest}
H.~Zhang, C.~Wu, Z.~Zhang, Y.~Zhu, H.~Lin, Z.~Zhang, Y.~Sun, T.~He, J.~Mueller,
  R.~Manmatha \emph{et~al.}, ``Resnest: Split-attention networks,'' \emph{arXiv
  preprint arXiv:2004.08955}, 2020.

\bibitem{lu2017expressive}
Z.~Lu, H.~Pu, F.~Wang, Z.~Hu, and L.~Wang, ``The expressive power of neural
  networks: a view from the width,'' in \emph{Nips}, 2017, pp. 6232--6240.

\bibitem{tan2019efficientnet}
M.~Tan and Q.~Le, ``Efficientnet: Rethinking model scaling for convolutional
  neural networks,'' in \emph{ICML}.\hskip 1em plus 0.5em minus 0.4em\relax
  PMLR, 2019, pp. 6105--6114.

\bibitem{zoph2016neural}
B.~Zoph and Q.~V. Le, ``Neural architecture search with reinforcement
  learning,'' \emph{ICLR}, 2017.

\bibitem{guo2020dmcp}
S.~Guo, Y.~Wang, Q.~Li, and J.~Yan, ``Dmcp: Differentiable markov channel
  pruning for neural networks,'' in \emph{CVPR}, 2020, pp. 1539--1547.

\bibitem{szege15}
C.~Szegedy, W.~Liu, Y.~Jia, P.~Sermanet, S.~Reed, D.~Anguelov, D.~Erhan,
  V.~Vanhoucke, and A.~Rabinovich, ``Going deeper with convolutions,'' in
  \emph{CVPR}, 2015, pp. 1--9.

\bibitem{szege16a}
C.~Szegedy, S.~Ioffe, V.~Vanhoucke, and A.~A. Alemi, ``Inception-v4,
  inception-resnet and the impact of residual connections on learning,'' in
  \emph{AAAI}, 2017, pp. 4278--4284.

\bibitem{he16}
K.~He, X.~Zhang, S.~Ren, and J.~Sun, ``Deep residual learning for image
  recognition,'' in \emph{CVPR}, 2016, pp. 770--778.

\bibitem{choll17}
F.~Chollet, ``Xception: Deep learning with depthwise separable convolutions,''
  in \emph{CVPR}, 2017, pp. 1251--1258.

\bibitem{nguyen2020wide}
T.~Nguyen, M.~Raghu, and S.~Kornblith, ``Do wide and deep networks learn the
  same things? {Uncovering} how neural network representations vary with width
  and depth,'' \emph{ICLR}, 2021.

\bibitem{zhu19}
Y.~Zhu, X.~Deng, and S.~Newsam, ``Fine-grained land use classification at the
  city scale using ground-level images,'' \emph{{{IEEE} Transactions on
  Multimedia}}, 2019.

\bibitem{cheng18}
X.~Cheng, J.~Lu, J.~Feng, B.~Yuan, and J.~Zhou, ``Scene recognition with
  objectness,'' \emph{{Pattern Recognition}}, vol.~74, pp. 474--487, 2018.

\bibitem{qiao2020urban}
Z.~Qiao, X.~Yuan, and M.~Elhoseny, ``Urban scene recognition via deep network
  integration,'' in \emph{International Conference on Urban Intelligence and
  Applications}.\hskip 1em plus 0.5em minus 0.4em\relax Springer, 2020, pp.
  135--149.

\bibitem{herra16}
L.~Herranz, S.~Jiang, and X.~Li, ``Scene recognition with cnns: objects, scales
  and dataset bias,'' in \emph{CVPR}, 2016.

\bibitem{wang17b}
Z.~Wang, L.~Wang, Y.~Wang, B.~Zhang, and Y.~Qiao, ``Weakly supervised
  patchnets: {Describing and aggregating local patches for scene
  recognition},'' \emph{{IEEE Transactions on Image Processing}}, vol.~26,
  no.~4, pp. 2028--2041, 2017.

\bibitem{zhou16}
B.~Zhou, A.~Khosla, A.~Lapedriza, A.~Torralba, and A.~Oliva, ``Places: An image
  database for deep scene understanding,'' \emph{Journal of Vision}, 2016.

\bibitem{xia19}
S.~Xia, J.~Zeng, L.~Leng, and X.~Fu, ``{WS-AM: Weakly Supervised Attention Map
  for Scene Recognition},'' \emph{Electronics}, vol.~8, no.~10, p. 1072, 2019.

\bibitem{selva17}
R.~Selvaraju, M.~Cogswell, A.~Das, R.~Vedantam, D.~Parikh, and D.~Batra,
  ``Visual explanations from deep networks via gradient-based localization,''
  in \emph{ICCV}, 2017, pp. 618--626.

\bibitem{seong19}
H.~Seong, J.~Hyun, H.~Chang, S.~Lee, S.~Woo, and E.~Kim, ``Scene recognition
  via object-to-scene class conversion: end-to-end training,'' in \emph{IJCNN},
  2019, pp. 1--6.

\bibitem{seong19a}
H.~Seong, J.~Hyun, and E.~Kim, ``Fosnet: An end-to-end trainable deep neural
  network for scene recognition,'' \emph{arXiv preprint arXiv:1907.07570},
  2019.

\bibitem{shi2019scene}
J.~Shi, H.~Zhu, S.~Yu, W.~Wu, and H.~Shi, ``Scene categorization model using
  deep visually sensitive features,'' \emph{IEEE Access}, vol.~7, pp.
  45\,230--45\,239, 2019.

\bibitem{wang2020deep}
C.~Wang, G.~Peng, and B.~De~Baets, ``Deep feature fusion through adaptive
  discriminative metric learning for scene recognition,'' \emph{Information
  Fusion}, vol.~63, pp. 1--12, 2020.

\bibitem{rehman2021scene}
A.~Rehman, S.~Saleem, U.~G. Khan, S.~Jabeen, and M.~O. Shafiq, ``Scene
  recognition by joint learning of dnn from bag of visual words and
  convolutional dct features,'' \emph{Applied Artificial Intelligence}, pp.
  1--19, 2021.

\bibitem{gupta2021visual}
S.~Gupta, K.~Sharma, D.~A. Dinesh, and V.~Thenkanidiyoor, ``Visual
  semantic-based representation learning using deep cnns for scene
  recognition,'' \emph{ACM Transactions on Multimedia Computing,
  Communications, and Applications}, vol.~17, no.~2, pp. 1--24, 2021.

\bibitem{qiao2021attention}
Z.~Qiao, X.~Yuan, C.~Zhuang, and A.~Meyarian, ``Attention pyramid module for
  scene recognition,'' in \emph{2020 25th International Conference on Pattern
  Recognition}.\hskip 1em plus 0.5em minus 0.4em\relax IEEE, 2021, pp.
  7521--7528.

\bibitem{yuan2021scale}
X.~Yuan, Z.~Qiao, and A.~Meyarian, ``Scale attentive network for scene
  recognition,'' \emph{Neurocomputing}, 2021.

\bibitem{zhou17}
B.~Zhou, A.~Lapedriza, A.~Khosla, A.~Oliva, and A.~Torralba, ``Places: A 10
  million image database for scene recognition,'' \emph{IEEE Transactions on
  Pattern Analysis and Machine Intelligence}, vol.~40, no.~6, pp. 1452--1464,
  2017.

\bibitem{fan2020analyzing}
Y.~Fan, Y.~Xian, M.~M. Losch, and B.~Schiele, ``Analyzing the dependency of
  convnets on spatial information,'' in \emph{DAGM German Conference on Pattern
  Recognition}.\hskip 1em plus 0.5em minus 0.4em\relax Springer, 2020, pp.
  101--115.

\bibitem{he2019bag}
T.~He, Z.~Zhang, H.~Zhang, Z.~Zhang, J.~Xie, and M.~Li, ``Bag of tricks for
  image classification with convolutional neural networks,'' in \emph{Computer
  Vision and Pattern Recognition}, 2019, pp. 558--567.

\bibitem{zhang2019making}
R.~Zhang, ``Making convolutional networks shift-invariant again,'' in
  \emph{International Conference on Machine Learning}.\hskip 1em plus 0.5em
  minus 0.4em\relax PMLR, 2019, pp. 7324--7334.

\end{thebibliography}

\vfill

\end{document}